\documentclass[11pt,a4paper]{article}
\usepackage[hyperref]{acl2021}
\usepackage{times}
\usepackage{latexsym}

\usepackage{graphicx}
\usepackage{subfigure}
\usepackage{booktabs}
\usepackage{multirow}
\usepackage{microtype}
\usepackage{amsthm,amsmath,amssymb}
\usepackage{mathrsfs}
\usepackage{colortbl}
\usepackage[normalem]{ulem}
\definecolor{mygray}{gray}{.95}
\renewcommand{\vec}[1]{\boldsymbol{#1}}
\aclfinalcopy 
\title{Semantic Representation for Dialogue Modeling}

\author{
 Xuefeng Bai$^{\spadesuit \heartsuit}$\hspace{0.5mm}, 
 Yulong Chen$^{\spadesuit \heartsuit}$\hspace{0.5mm}, 
 Linfeng Song$^{\clubsuit}$\hspace{0.5mm}, 
 Yue Zhang$^{\heartsuit \diamondsuit}$\hspace{0.2mm}\hspace{1.5mm} \\
 $^\spadesuit$ Zhejiang University, China\\
 $^\heartsuit$ School of Engineering, Westlake University, China\\
 $^\clubsuit$ Tencent AI Lab, Bellevue, WA, USA\\
 $^\diamondsuit$ Institute of Advanced Technology, Westlake Institute for Advanced Study, China
}

\date{}

\begin{document}
\maketitle
\begin{abstract}
Although neural models have achieved competitive results in dialogue systems, they have shown limited ability in representing core semantics, such as ignoring important entities.
To this end, we exploit Abstract Meaning Representation (AMR) to help dialogue modeling. 
Compared with the textual input, AMR explicitly provides core semantic knowledge and reduces data sparsity.
We develop an algorithm to construct dialogue-level AMR graphs from sentence-level AMRs and explore two ways to incorporate AMRs into dialogue systems.
Experimental results on both dialogue understanding and response generation tasks show the superiority of our model.
To our knowledge, we are the first to leverage a formal semantic representation into neural dialogue modeling.
\end{abstract}

\section{Introduction}
Dialogue systems have received increasing research attention \cite{wen2015semantically,serban2016hierarchical,bao-etal-2020-plato}, with much recent work focusing on social chats ~\citep{ritter-etal-2011-data,li-etal-2017-dailydialog} and task-oriented dialogues~\citep{wen-etal-2017-network,dinan2018wizard}.
There are two salient subtasks in dialogue modeling, namely dialogue understanding~\citep{ChoiHIYYCLZ18,ReddyCM19,yu-2020-dialogue} and response generation~\citep{li-etal-2017-dailydialog,budzianowski2018multiwoz}.
The former refers to understanding of semantic and discourse details in a dialogue history, and the latter concerns making a fluent, novel and coherent utterance.
\begin{figure}[t!]
	\centering
	\includegraphics[width=0.9\columnwidth]{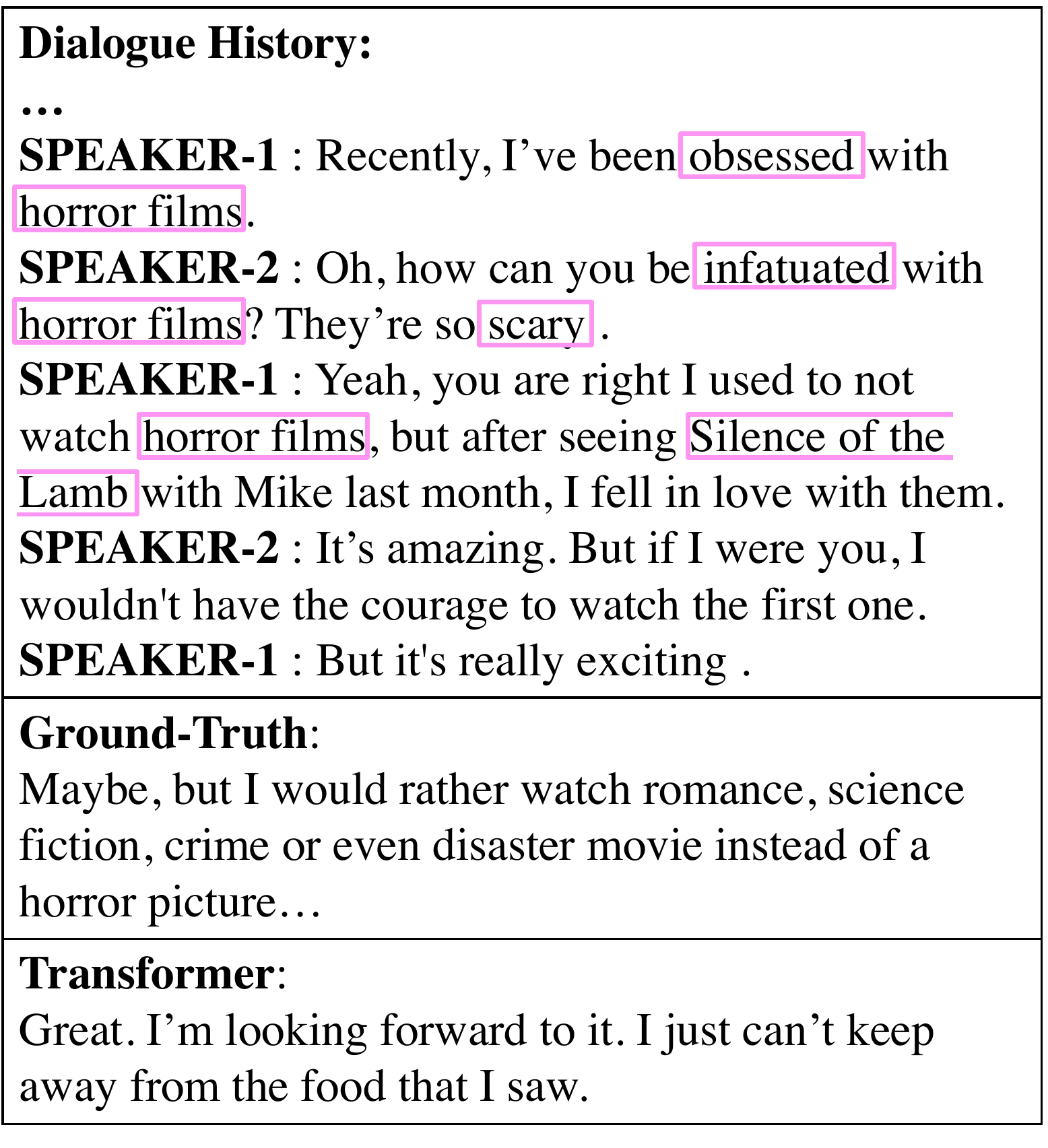}
	\caption{A conversation from DailyDialog. Some important contents are marked with squares.}
	\label{fig:example}
\end{figure}

The current state-of-the-art methods employ neural networks and end-to-end training \citep{Sutskever14,BahdanauCB14} for dialogue modeling.
For instance, sequence-to-sequence models have been used to encode a dialogue history, before directly synthesizing the next utterance \citep{VinyalsL15, wen-etal-2017-network,bao-etal-2020-plato}. 
Despite giving strong empirical results, neural models can suffer from spurious feature associations in their neural semantic representation \cite{poliak-etal-2018-hypothesis,kaushik2020learning}, which can lead to weak robustness, inducing irrelevant dialogue states \cite{Xu2014SLU,sharma-etal-2019-improving,rastogi-etal-2019-scaling} and generating unfaithful or irrelevant text~\cite{maynez2020faithfulness,niu2020avgout}.
As shown in Figure~\ref{fig:example}, the baseline Transformer model pays attention to the word ``\textit{lamb}'' but ignores its surrounding context, which has important contents (marked with squares) that indicate its true meaning, thereby giving an irrelevant response that is related to food.
Intuitively, such issues can be alleviated by having a structural representation of semantic information, which treats entities as nodes and builds structural relations between nodes, making it easy to find the most salient context. 
Explicit structures are also more interpretable compared to neural representation and have been shown useful for information extraction~\cite{StrubellVAWM18,sun-etal-2019-aspect,Li20LocalAttention,bai21syntax,SachanZQH21}, summarization \cite{liu-etal-2015-toward,hardy-vlachos-2018-guided,liao2018abstract} and machine translation \cite{marcheggiani-etal-2018-exploiting,song2019semantic}.

\begin{figure*}[!t]
	\centering
	\includegraphics[width=0.9\textwidth]{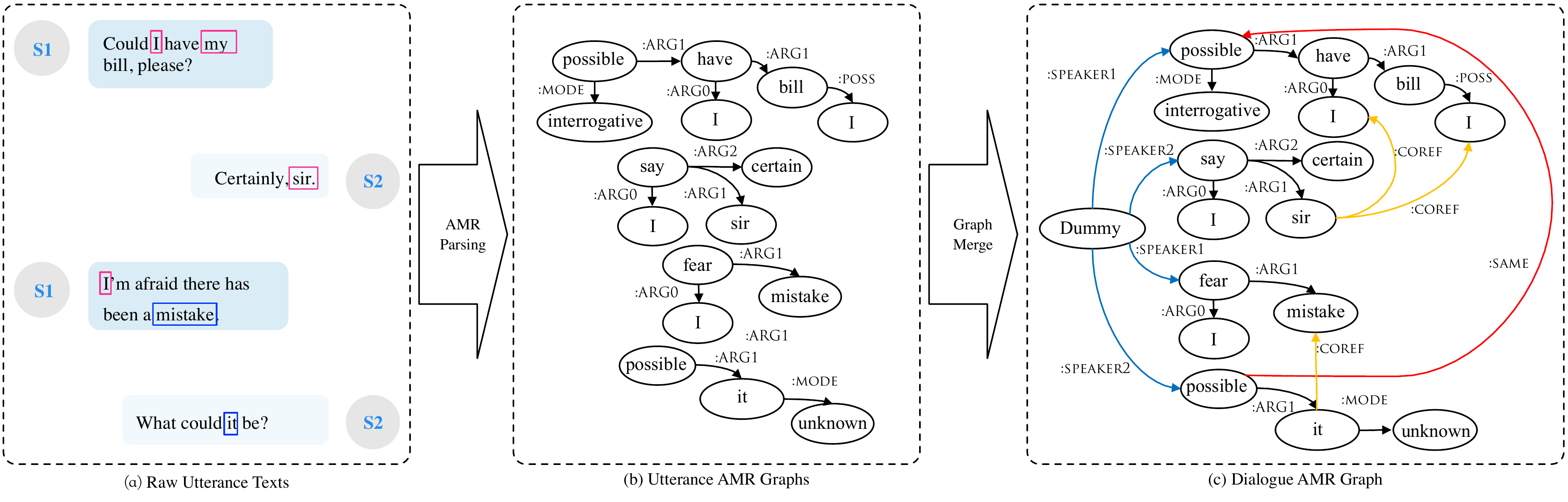}
	\caption{Dialogue AMR graph construction process. Step 1: parse raw-text utterance into utterance AMR graphs; Step 2: connect utterance AMR graphs into a dialogue AMR graph.}
	\label{amr_graphs}
\end{figure*}

We explore AMR~\cite{banarescu2013abstract} as a semantic representation for dialogue histories in order to better represent conversations.
As shown in the central block of Figure~\ref{amr_graphs}, AMR is one type of sentential semantic representations, which models a sentence using a rooted directed acyclic graph, highlighting its main concepts (\emph{e.g.} ``\textit{mistake}'') and semantic relations (\emph{e.g.}, ``\textit{ARG0}''\footnote{Please refer to PropBank~\cite{KingsburyP02,PalmerKG05} for more details.}), while abstracting away function words.
It can thus potentially offer core concepts and explicit structures needed for aggregating the main content in dialogue.
In addition, AMR can also be useful for reducing the negative influence of variances in surface forms with the same meaning, which adds to data sparsity. 
Existing work on AMR parsing focuses on the sentence level. 
However, as the left block of Figure~\ref{amr_graphs} shows, the semantic structure of a dialogue history can consist of rich cross-utterance co-reference links (marked with squares) and multiple speaker interactions.
To this end, we propose an algorithm to automatically derive dialogue-level AMRs from utterance-level AMRs, by adding cross-utterance links that indicate speakers, identical mentions and co-reference links. 
One example is shown in the right block of Figure~\ref{amr_graphs}, where newly added edges are in color.
We consider two main approaches of making use of such dialogue-level AMR structures. 
For the first method, we merge an AMR with tokens in its corresponding sentence via AMR-to-text alignments, before encoding the resulting structure using a graph Transformer~\cite{zhu2019modeling}. 
For the second method, we separately encode an AMR and its corresponding sentence, before leveraging both representations via feature fusion~\cite{Mangai10} or dual attention~\cite{calixto2017doubly}.

We verify the effectiveness of the proposed framework on a dialogue relation extraction task~\citep{yu-2020-dialogue} and a response generation task~\citep{li-etal-2017-dailydialog}.
Experimental results show that the proposed framework outperforms previous methods~\citep{vaswani2017attention,bao-etal-2020-plato,yu-2020-dialogue}, achieving the new state-of-the-art results on both benchmarks. 
Deep analysis and human evaluation suggest that semantic information introduced by AMR can help our model to better understand long dialogues and improve the coherence of dialogue generation.
One more advantage is that AMR is helpful to enhance the robustness and has a potential to improve the interpretability of neural models.
To our knowledge, this is the first attempt to leverage the AMR semantic representation into neural networks for dialogue understanding and generation.
Our code is available at~\url{https://github.com/muyeby/AMR-Dialogue}.

\section{Constructing Dialogue AMRs}
\label{sec:dialogue-AMR}


Figure~\ref{amr_graphs} illustrates our method for constructing a dialogue-level AMR graph from multiple utterance-level AMRs.
Given a dialogue consisting multiple utterances, we adopt a pretrained AMR parser~\citep{cai-lam-2020-amr} to obtain an AMR graph for each utterance.
For utterances containing multiple sentences, we parse them into multiple AMR graphs, and mark them belonging to the same utterance.
We construct each dialogue AMR graph by making connections between utterance AMRs.
In particular, we take three strategies according to speaker, identical concept and co-reference information.

\paragraph{Speaker} We add a dummy node and connect it to all root nodes of utterance AMRs.
We add speaker tags (\emph{e.g.}, \textsc{speaker1} and \textsc{speaker2}) to the edges to distinguish different speakers. 
The dummy node ensures that all utterance AMRs are connected so that information can be exchanged during graph encoding.
Besides, it serves as the global root node to represent the whole dialogue.

\paragraph{Identical Concept} There can be identical mentions in different utterances (\emph{e.g.} ``\textit{possible}'' in the first and the forth utterances in Figure~\ref{amr_graphs}), resulting in repeated concept nodes in utterance AMRs.
We connect nodes corresponding to the same \textbf{non-pronoun} concepts by edges labeled with \textsc{same}\footnote{Compared with co-reference, \textit{identical concept} relations can connect different words which share the same meaning \emph{e.g.}$\left\langle \text{could}, \text{might} \right\rangle, \left\langle \text{fear}, \text{afraid} \right\rangle$.}. 
This type of connection can further enhance cross-sentence information exchange.

\paragraph{Inter-sentence Co-reference} A major challenge for dialogues understanding is posed by pronouns, which are frequent in conversations~\citep{grosz-etal-1995-centering, newman2008gender, quan2019gecor}. 
We conduct co-reference resolution  on dialogue text using an off-to-shelf model\footnote{https://github.com/huggingface/neuralcoref} in order to identify concept nodes in utterance AMRs that refer to the same entity.
For example, in Figure~\ref{amr_graphs}, ``\textit{I}'' in the first utterance, and ``\textit{sir}'' in the second utterance refer to the same entity, \textsc{speakr1}. 
We add edges labeled with \textsc{coref} between them, starting from \textit{later} nodes to \textit{earlier} nodes (\textit{later} and \textit{earlier} here refer to the temporal order of ongoing conversation), to indicate their relation\footnote{For simplicity, we omit the coreference links between the second and third utterance for display.}.


\section{Baseline System}
\label{sec:stdTransformer}
We adopt a standard Transformer~\cite{vaswani2017attention} for dialogue history encoding. 
Typically, a Transformer encoder consists of $L$ layers, taking a sequence of tokens (i.e., dialogue history) $\mathcal{S} = \{{w}_1, {w}_2, ..., {w}_N\}$, where $w_i$ is the $i$-th token and $N$ is the sequence length, as input and produces vectorized word representations $\{h_1^l, h_2^l, ..., h_N^l\}$ iteratively, $l \in [1,...,L]$. 
Overall, a Transformer encoder can be written as:
\begin{equation}
	\label{eq:seqenc}
	H = \texttt{SeqEncoder}(\texttt{emb}(\mathcal{S})),
\end{equation}
where $H = \{ h_1^L, h_2^L, ...,h_n^L\}$, and \texttt{emb} denotes a function that maps a sequence of tokens into the corresponding embeddings.
Each Transformer layer consists of two sub-layers: a self-attention sub-layer and a position-wise feed forward network.
The former calculates a set of attention scores:
\begin{equation}
    \alpha_{ij} = \texttt{Attn}(h_i, h_j).
    \label{eq:dotattention}
\end{equation}
which are used to update the hidden state of $w_i$:
\begin{equation}
	h^{l}_{i} = \sum\nolimits_{j = 1}^N \alpha_{ij}(W^{V}{h}^{l-1}_{j}),
	\label{eq:update}
\end{equation}
where $W^V$ is a parameter matrix.

The position-wise feed-forward (\texttt{FFN}) layer consists of two  linear transformations:
\begin{equation}
	\texttt{FFN}(h) = W_2\texttt{ReLU}(W_1h + b_1) + b_2,
	\label{eq:FFN}
\end{equation}
where $W_1, W_2, b_1, b_2$ are model parameters.

\subsection{Dialogue Understanding Task}
We take the dialogue relation extraction task~\cite{yu-2020-dialogue} as an example. 
Given a dialogue history $\mathcal{S}$ and an argument (or entity) pair ($a_1$, $a_2$), the goal is to predict the corresponding relation type $r \in \mathcal{R}$ between $a_1$ and $a_2$.

We follow a previous dialogue relation extraction model~\cite{Chen20Dialogue} to feed the hidden states of $a_1$ and $a_2$ (denoted as $h_{a_1}, h_{a_2}$) into a classifier to obtain the probability of each relation types:
\begin{equation}
    \label{eq:relationprob}
	P_{rel} = \texttt{softmax}(W_3[h_{a_1};h_{a_2}] +b_3),
\end{equation}
where 
$W_3$ and $b_3$ are model parameters. 
The \textit{k}-th value of $P_{rel}$ is the conditional probability of \textit{k}-th relation in $\mathcal{R}$.

Given a training instance $ \left\langle \mathcal{S}, a_1, a_2, r\right\rangle$, the local loss is:
\begin{equation}
	\ell = -logP({r}| \mathcal{S}, a_1, a_2;\theta),
\end{equation}
where $\theta$ denotes the set of model parameters.
In practice, we use BERT \cite{devlin-etal-2019-bert} for calculating $h_{a_1}$ and $h_{a_2}$, which can be regarded as pre-trained initialization of the Transformer encoder.

\begin{figure*}[t!]
	\centering 
	\subfigure[ ]{\includegraphics[width=0.3\hsize]{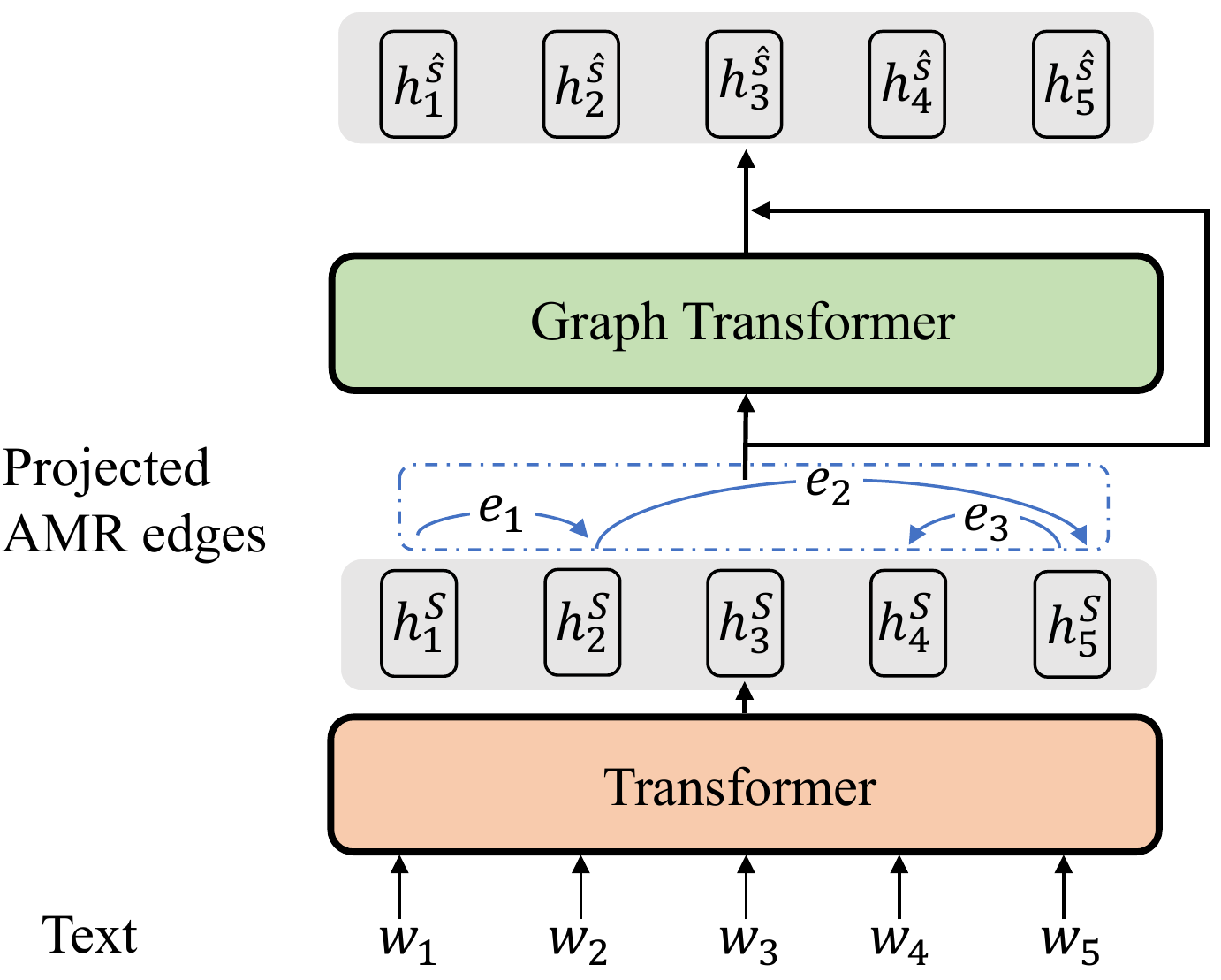}\label{fig:model1}} \hspace{0.15in}
	\subfigure[ ]{\includegraphics[width=0.3\hsize]{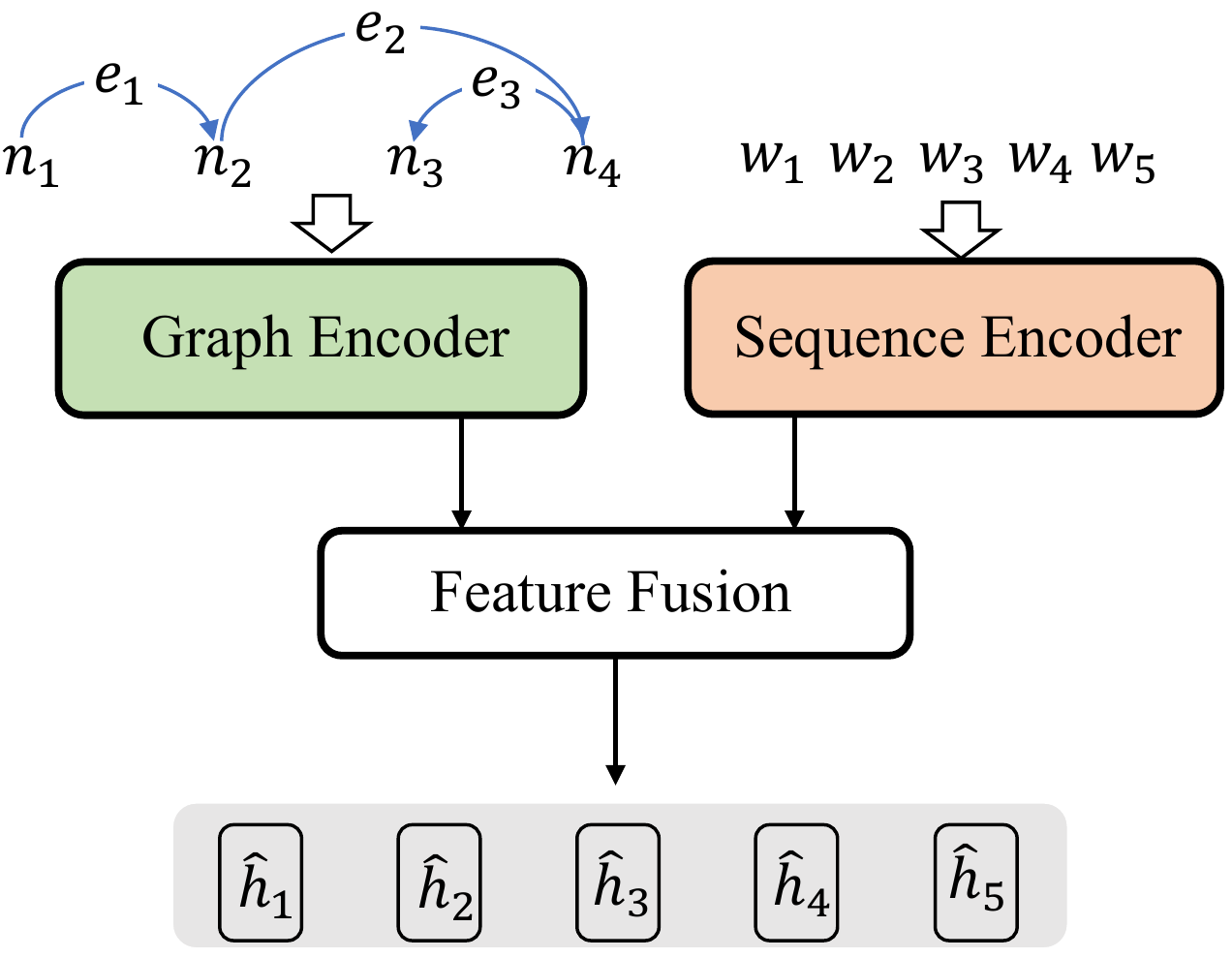}\label{fig:model2}} \hspace{0.15in}
	\subfigure[ ]{\includegraphics[width=0.3\hsize]{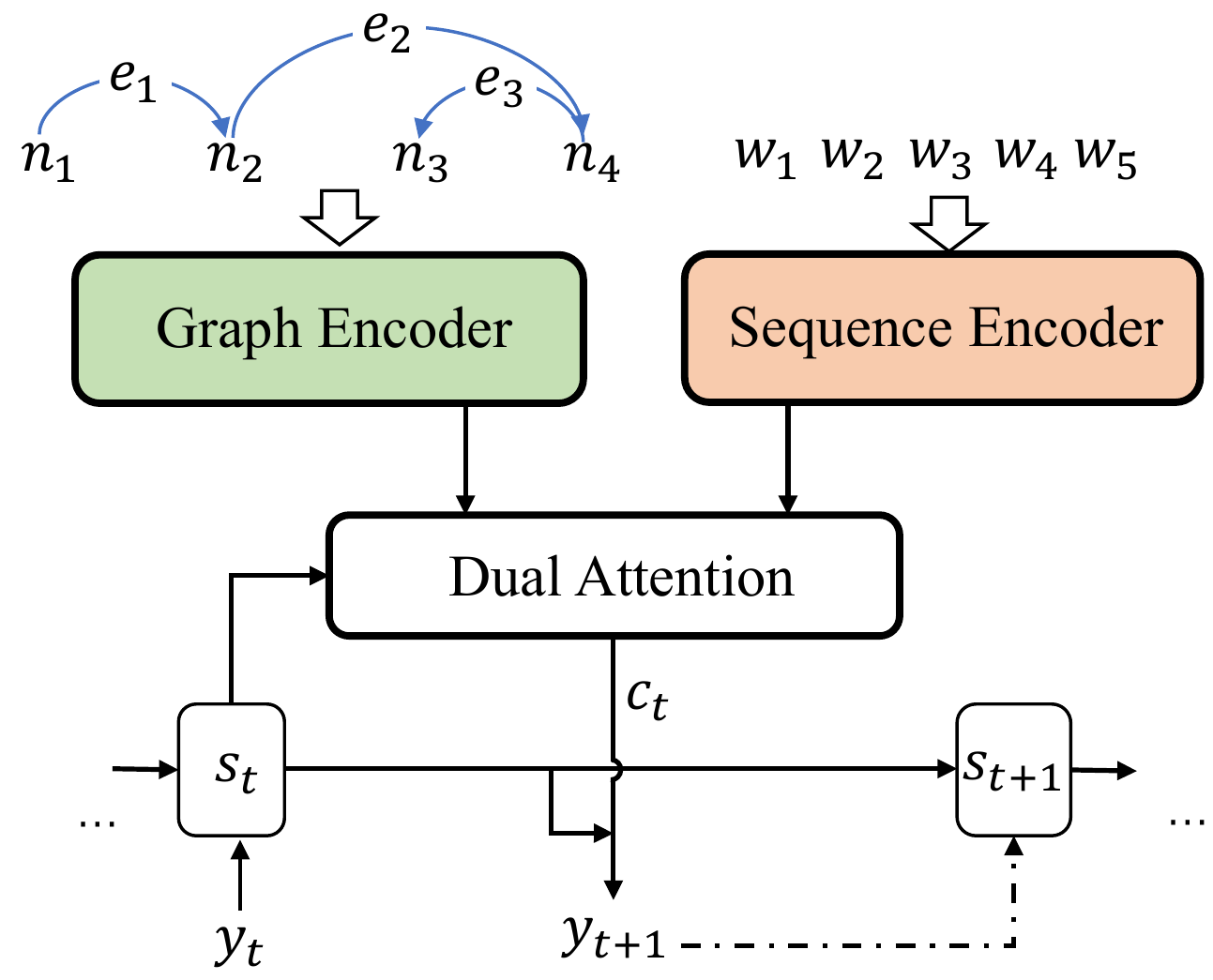}\label{fig:model3}}\\
	\caption{AMR for dialogue modeling. (a) Using AMR to enrich text representation. (b,c) Using AMR independently.}
	\label{fig:model}
\end{figure*}
\subsection{Dialogue Response Generation Task}
Given a dialogue history $\mathcal{S}$, we use a standard auto-regressive Transformer decoder~\cite{vaswani2017attention} to generate a response $\mathcal{Y}=\{y_1,y_2,...,y_{|\mathcal{Y}|}\}$.
At time step $t$, the previous output word $y_{t-1}$ is firstly transformed into a hidden state $s_t$ by a self-attention layer as Equations~\ref{eq:dotattention} and~\ref{eq:update}.
Then an encoder-decoder attention mechanism is applied to obtain a context vector from encoder output hidden states$\{h_1^L, h_2^L, \dots, h_N^L\}$:
\begin{equation}
	\label{eq:crossattn}
	\begin{split}
		\hat{\alpha}_{ti} &= \texttt{Attn}(s_t, h^{L}_{i}), \\
		c_{t} &= \sum\nolimits_{i=1}^N \hat{\alpha}_{ti} h^{L}_{i}, \\
	\end{split}
\end{equation}

The obtained context vector $c_t$ is then used to calculate the output probability distribution for the next word $y_{t}$ over the target
vocabulary\footnote{Similar to the encoder, there is also multi-head attention, a position-wise feed-forward layer and residual connections, which we omit in the equations.}:
\begin{equation}
	\label{eq:tgtprob}
	P_{voc} = \texttt{softmax}(W_4c_t+b_4), \\
\end{equation}
where $W_4, b_4$ are trainable model parameters. The \textit{k}-th value of $P_{voc}$ is the conditional probability of \textit{k}-th word in vocabulary given a dialogue.

Given a dialogue history-response pair $\{\mathcal{S}, \mathcal{Y}\}$, the model minimizes a cross-entropy loss:
\begin{equation}
	\label{eq:drg-obj}
	\ell = -\sum_{t=1}^{|Y|}log P_{voc}(y_{t}|y_{t-1}, ...,y_1, \mathcal{S}; \theta) , \\
\end{equation}
where $\theta$ denotes all model parameters.

\section{Proposed Model}
Our model takes a dialogue history $\mathcal{S}$ and the corresponding dialogue AMR as input. 
Formally, an AMR is a directed acyclic graph $\mathcal{G} = \left\langle\mathcal{V}, \mathcal{E}\right\rangle$, where $\mathcal{V}$ denotes a set of nodes (i.e. AMR concepts) and $\mathcal{E}$ (i.e. AMR relations) denotes a set of labeled edges.
An edge can be further represented by a triple $\left\langle n_i, r_{ij}, n_j\right\rangle$, meaning that the edge is from node $n_i$ to $n_j$ with label $r_{ij}$.
We consider two main ways of making use of dialogue-level AMRs. 
The first method (Figure~\ref{fig:model1}) uses AMR semantic relations to enrich a textual representation of the dialogue history. 
We project AMR nodes onto the corresponding tokens, extending Transformer by encoding semantic relations between words.
For the second approach, we separately encode an AMR and its sentence, and use either feature fusion (Figure~\ref{fig:model2}) or dual attention (Figure~\ref{fig:model3}) to incorporate their embeddings.

\subsection{Graph Encoding}
\label{sec:graphTransformer}
We adopt a Graph Transformer~\cite{zhu2019modeling} to encode an AMR graph,
which extends the standard Transformer~\cite{vaswani2017attention} for modeling structural input. 
A $L$-layer graph Transformer takes a set of node embeddings $\{\vec{n}_1, \vec{n}_2, ..., \vec{n}_M\}$ and a set of edge embeddings $\{\vec{r}_{ij}| i \in [1,...,M], j \in [1,...,M]\}$ as input\footnote{If there is no relation between $n_i$ and $n_j$, $r_{ij}$=``\texttt{None}''} and produces more abstract node features $\{h_1^l, h_2^l, ..., h_M^l\}$ iteratively, where $l \in [1,...,L]$. 
The key difference between a graph Transformer and a standard Transformer is the graph attention layer. 
Compared with self-attention layer (Equation~\ref{eq:dotattention}), the graph attention layer explicitly considers graph edges when updating node hidden states.
For example, give an edge $\left\langle n_i, r_{ij}, n_j\right\rangle$, the attention score $\hat{\alpha}_{ij}$ is calculated as:
\begin{equation}
	\begin{split}
		\hat{\alpha}_{ij} &= \frac{\exp(\hat{e}_{ij})}{\sum\nolimits_{m = 1}^M \exp{(\hat{e}_{im})}}, \\
		\hat{e}_{ij} &= \frac{(W^Qh_{i}^{l-1})^{T}(W^{K}h_{j}^{l-1}+W^R\vec{r}_{ij})}{\sqrt{d}},\\
	\end{split}
	\label{eq:relation-dotattention}
\end{equation}
where $W^R$ is a transformation matrix, $\vec{r}_{ij}$ is the embedding of relation $r_{ij}$, $d$ is hidden state size, and $\{h_1^0,h_2^0,...,h_M^0 \} = \{\vec{n}_1, \vec{n}_2, ..., \vec{n}_M\}$.
The hidden state of $n_i$ is then updated as:
\begin{equation}
	h^{l}_{i} = \sum\nolimits_{j = 1}^M \alpha_{ij}(W^{V}{h}^{l-1}_{j} + W^R\vec{r}_{ij}),
	\label{eq:relation-update}
\end{equation}
where $W^V$ is a parameter matrix.
Overall, given an input AMR graph $\mathcal{G} = \left\langle\mathcal{V}, \mathcal{E}\right\rangle$, the graph Transformer encoder can be written as
\begin{equation}
	\label{eq:graphenc}
	H = \texttt{GraphEncoder}(\texttt{emb}(\mathcal{V}), \texttt{emb}(\mathcal{E})),
\end{equation}
where $H = \{ h_1^L, h_2^L, ...,h_M^L\}$ denotes top-layer graph encoder hidden states.

\subsection{Enriching Text Representation}
\label{sec:refine}
We first use the JAMR aligner~\cite{flanigan-etal-2014-discriminative} to obtain a node-to-word alignment, then adopt the alignment to project the AMR edges onto text with following rules:
\begin{equation}
	\hat{r}_{ij}=\left\{
	\begin{aligned}
		r_{i'j'} &, &\text{if}~\mathcal{A}(n_{i'}) = w_{i}, \mathcal{A}(n_{j'}) = w_{j}, \\
		\texttt{Self} &,& \text{if}~i=j, \\
		\texttt{None} &,& \text{otherwise} \text{,}
	\end{aligned}
	\right.
	\label{eq:projectededge}
\end{equation}
where $\mathcal{A}$ is a one-to-$K$ alignment ($K\in [0,\dots, N]$). 
In this way, we obtain a \textit{projected} graph $\mathcal{G'} = \left\langle\mathcal{V'}, \mathcal{E'}\right\rangle$, where $\mathcal{V'}$ represents the set of input words $\{w_1, w_2, ..., w_N\}$ and $\mathcal{E'}$ denotes a set of \textit{word-to-word} semantic relations.

Inspired by previous work on AMR graph modeling~\cite{guo-etal-2019-attention,song-etal-2019-leveraging,sun-etal-2019-aspect}, we adopt a hierarchical encoder that stacks a sequence encoder and a graph encoder. 
A sequence encoder (\texttt{SeqEncoder}) transforms a dialogue history into a set of hidden states:
\begin{equation}
	H^S  = \texttt{SeqEncoder}(\texttt{emb}(\mathcal{S})).
	\label{eq:textmemory}
\end{equation}

A graph encoder incorporates the \textit{projected} relations features into $H^S$:
\begin{equation}
	H^{\hat{S}} = \texttt{GraphEncoder}(H^S, \texttt{emb}(\mathcal{E'})),
	\label{eq:graphadapter}
\end{equation}

In addition, we add a residual connection between graph adapter and sequence encoder to fuse word representations before and after refinement (as shown in Figure~\ref{fig:model2}):
\begin{equation}
	H^{F} = \texttt{LayerNorm}(H^S + H^{\hat{S}}).
	\label{eq:fusedrepresentation}
\end{equation}
where \texttt{LayerNorm} denotes the layer normalization~\cite{BaKH16}.
We name the hierarchical encoder as \texttt{Hier}, which can be used for both dialogue understanding and dialogue response generation.

\subsection{Leveraging both Text and Structure Cues}
\label{sec:exmem}
We consider integrating both text cues and AMR structure cues for dialogue understanding and response generation, using a dual-encoder network.
First, a sequence encoder is used to transform a dialogue history $\mathcal{S}$ into a \textit{text memory} (denoted as $H^S = \{h^{S}_{1},h^{S}_{2},...,h^{S}_{N}\}$) using Equation~\ref{eq:seqenc}.
Second, the AMR graph $\mathcal{G}$ is encoded into \textit{graph memory} (denoted as $H^G = \{h^{G}_{1},h^{G}_{2},...,h^{G}_{M}\}$) by a graph Transformer encoder using Equation~\ref{eq:graphenc}.

For dialogue understanding (Figure~\ref{fig:model2}) and dialogue response generation (Figure~\ref{fig:model3}), slightly different methods of feature integration are used due to their different nature of outputs.

\noindent\textbf{Dialogue Understanding}.
Similar to Section~\ref{sec:refine}, we first use the JAMR aligner to obtain a node-to-word alignment~$\mathcal{A}$.
Then we fuse the word and AMR node representations as follows:
\begin{equation}
	\hat{h}_{i}=\left\{
	\begin{aligned}
		f(h_i^S, h^G_j) &, & \text{if}~\exists j,~\mathcal{A}(n_j) = w_i, \\
		f(h_i^S, h_{\emptyset}) &,& \text{otherwise},
	\end{aligned}
	\right.
	\label{eq:projectednode}
\end{equation}
where $h_{\emptyset}$ is the vector representation of the dummy node (see Figure~\ref{amr_graphs}), $f$ is defined as:
\begin{equation}
	h = \texttt{LayerNorm}(h_1 + h_2).
\end{equation}
The fused word representations are then fed into a classifier for relation prediction (Equation~\ref{eq:relationprob}).

\noindent\textbf{Dialogue Response Generation}.
We replace the standard encoder-decoder attention (Equation~\ref{eq:crossattn}) with a dual-attention mechanism~\cite{song2019semantic}. 
In particular, given a decoder hidden state $s_t$ at time step $t$, the dual-attention mechanism calculates a graph context vector $c_{t}^S$ and a text context vector $c_{t}^G$, simultaneously:
\begin{table*}[!t]
	\centering
	\small
	\begin{tabular}{lcccccccc}
		\toprule
		\multirow{3}{*}{\textbf{Model}}& \multicolumn{4}{c}{\textbf{data-v1}} &\multicolumn{4}{c}{\textbf{data-v2}} \\
		\cmidrule(lr){2-5} \cmidrule(lr){6-9}
		&\multicolumn{2}{c}{\textbf{dev}} &\multicolumn{2}{c}{\textbf{test}}
		&\multicolumn{2}{c}{\textbf{dev}} 
		&\multicolumn{2}{c}{\textbf{test}} \\
		&F1$(\delta)$ &F$1_c(\delta)$ &F1$(\delta)$ &F1$_c(\delta)$
		&F1$(\delta)$ &F$1_c(\delta)$ &F1$(\delta)$ &F1$_c(\delta)$\\
		\midrule
		AGGCN$^\dagger$ &46.6(-) &40.5(-) &46.2(-) &39.5 (-) &- &- &- &-\\
		LSR$^\dagger$ &44.5(-) &- &44.4(-) &- &- &- &- &- \\
		DHGAT$^\dagger$ &57.7(-) &52.7(-) &56.1(-) &50.7(-) &- &- &- &- \\
		BERT &60.6(1.2) &55.4(0.9) &58.5(2.0) &53.2(1.6) &59.4 (0.7) &54.7(0.8) &57.9(1.0)	&53.1(0.7)\\
		{BERT$_s$} &{63.0(1.5)} &{57.3(1.2)} &{61.2(0.9)} &{55.4(0.9)} &62.2(1.3) &57.0(1.0) &59.5(2.1) &54.2(1.4) \\
		\midrule
		\rowcolor{mygray}
		BERT$_{c}$ &66.8(0.9) &60.9(1.0) &66.1(1.1) &60.2(0.8) &66.2(0.9) &60.5(1.1) &65.1(0.8) &59.8(1.2) \\
		\rowcolor{mygray}
		\texttt{Hier} &68.2(0.8) &\textbf{62.2}(0.7) &67.0(0.9) &61.3(0.6) &68.0(0.6) &62.2(0.4) &66.7(0.3) &61.0(0.4) \\
		\rowcolor{mygray}
		\texttt{Dual} &\textbf{68.3}(0.6) &\textbf{62.2}(0.2) &\textbf{67.3}(0.4) &\textbf{61.4}(0.2) &\textbf{68.2}(0.5) &\textbf{62.3}(0.4) &\textbf{67.1}(0.4) &\textbf{61.1}(0.5) \\
		\bottomrule
	\end{tabular}
	\caption{Performance on DialogRE, where $\delta$ denotes the standard deviation computed from 5 runs, and $\dagger$ indicates results reported by~\citet{Chen20Dialogue}.}
	\label{tab:mainRE}
\end{table*}
\begin{equation}
	\label{eq:text_graphcontext}
	\begin{split}
		&\hat{\alpha}_{ti} = \texttt{Attn}(s_t, h^{S}_{i}), \\
		&\hat{\alpha}_{tj} = \texttt{Attn}(s_t, h^{G}_{j}), \\
		&c_{t}^S = \sum\nolimits_{i=1}^N \hat{\alpha}_{ti} h^{S}_{i}, \\
		&c_{t}^G =  \sum\nolimits_{j=1}^M \hat{\alpha}_{tj} h^{G}_{j}, \\
	\end{split}
\end{equation}
and the final context vector $\hat{c}_t$ is calculated as:
\begin{equation}
	c_t  = W^c[c_{t}^S;c_{t}^G] + b^c,
	\label{eq:finalcontext}
\end{equation}
where $W^c$ and $b^c$ are model parameters.

We name the dual-encoder model as \texttt{Dual}.

\section{Dialogue Understanding Experiments}
We evaluate our model on DialogRE~\cite{yu-2020-dialogue}, which contains totally 1,788 dialogues, 10,168 relational triples and 36 relation types in total. 
On average, a dialogue in DialogRE contains 4.5 relational triples and 12.9 turns.
We report experimental results on both original (v1) and updated (v2) English version.\footnote{\url{https://dataset.org/dialogre/}}


\subsection{Settings}

We adopt the same input format and hyperparameter settings as~\citet{yu-2020-dialogue} for the proposed model and baselines.
In particular, the input sequence is constructed as $\texttt{[CLS]} d \texttt{[SEP]} a_1 \texttt{[SEP]} a_2 \texttt{[SEP]}$, where $d$ denotes the dialogue, and $a_1$ and $a_2$ are the two associated arguments.
In the BERT model of~\citet{yu-2020-dialogue}, only the hidden state of the \texttt{[CLS]} token is fed into a classifier for prediction, while our baseline (BERT$_c$) additionally takes the hidden states of $a_1$ and $a_2$.
All hyperparameters are selected by prediction accuracy on validation dataset (See Table~\ref{tab:hyperparameters} for detailed hyperparameters).

\noindent\textbf{Metrics} Following previous work on DialogRE, we report macro F$1$ score on relations in both the standard (F$1$) and conversational settings (F$1_c$; \citealp{yu-2020-dialogue}). 
F$1_c$ is computed over the first few turns of a dialogue where two arguments are first mentioned.

\subsection{Main Results}
Table~\ref{tab:mainRE} shows the results of different systems on DialogRE. 
We compare the proposed model with two BERT-based approches, BERT and BERT$_s$. 
Based on BERT, BERT$_s$~\cite{yu-2020-dialogue} highlights speaker information by replacing speaker arguments with special tokens.
For completeness, we also include recent methods, such as 
AGGCN~\cite{guo-etal-2019-attention}, LSR~\cite{nan-etal-2020-reasoning} and DHGAT~\cite{Chen20Dialogue}.
BERT$_{c}$ and \texttt{Hier}, \texttt{Dual} represent our baseline and the proposed models, respectively.

By incorporating speaker information, BERT$_s$ gives the best performance among the previous system.
Our BERT$_{c}$ baseline outperforms BERT$_s$ by a large margin, as BERT$_{c}$ additionally considers argument representations for classification.
\texttt{Hier} significantly ($p<0.01$)\footnote{We use pair-wised $t$-test.} outperforms BERT$_{c}$ in all settings, with 1.4 points of improvement in terms of F1 score on average.
A similar trend is observed under F1$_c$.
This shows that semantic information in AMR is beneficial to dialogue relation extraction, since AMR highlights core entities and semantic relations between them.
\texttt{Dual} obtains slightly better results than \texttt{Hier}, which shows effect of separately encoding a semantic structure.

Finally, the standard deviation values of both \texttt{Dual} and \texttt{Hier} are lower than the baselines. 
This indicates that our approaches are more robust regarding model initialization.
\begin{figure}
    \centering
    \includegraphics[width=0.95\columnwidth]{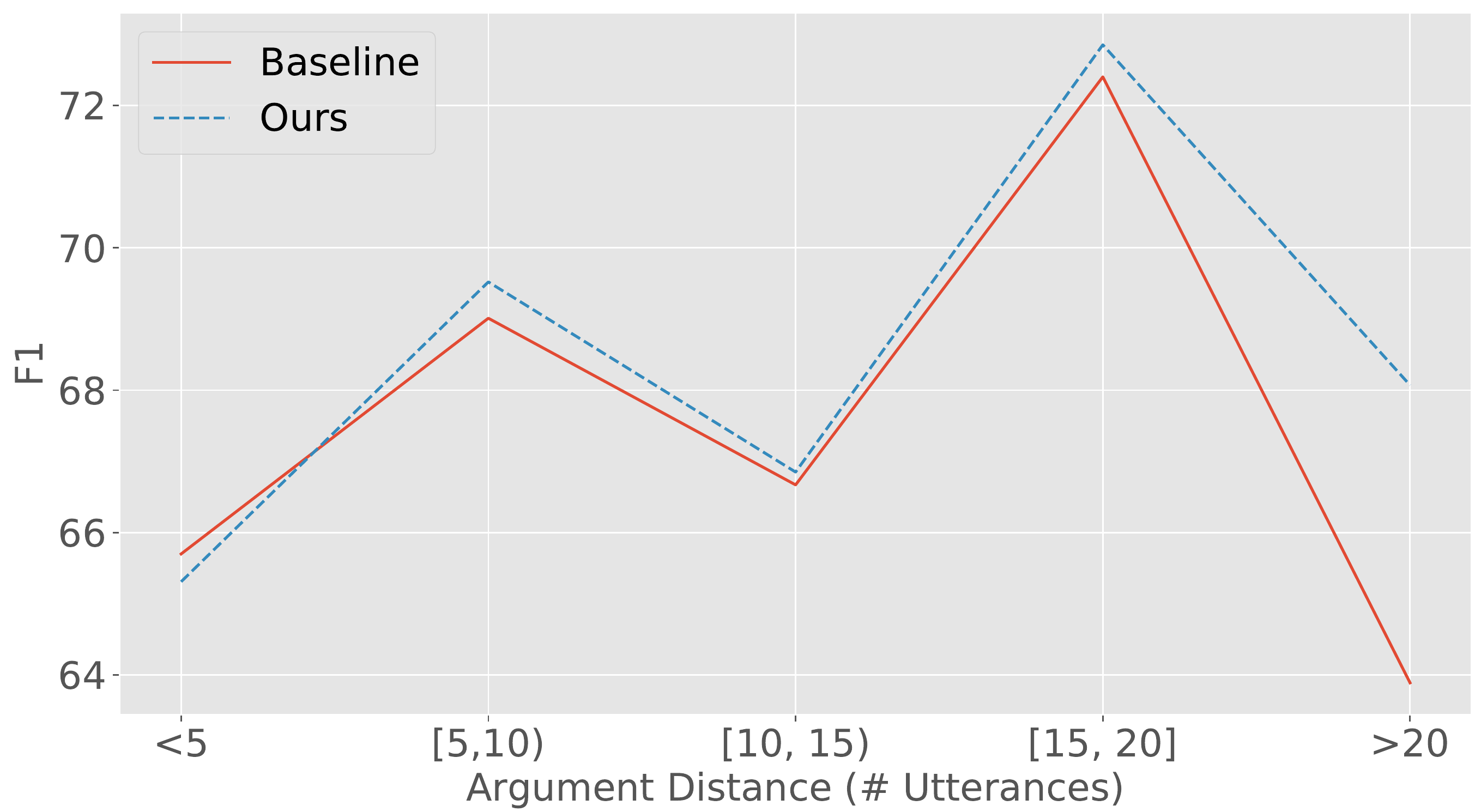}
    \caption{The performance of BERT$_c$ (Baseline) and \texttt{Dual} (Ours) regarding argument distances.}
    \label{fig:dis-utter}
\end{figure}
\subsection{Impact of Argument Distance}
We split the dialogues of the DialogRE (v2) devset into five groups by the utterance-based distance between two arguments.
As shown in Figure~\ref{fig:dis-utter}, \texttt{Dual} gives better results than BERT$_c$ except when the argument distance is less than 5.
In particular, \texttt{Dual} surpasses BERT$_c$ by a large margin when the arguments distance is greater than 20.
The comparison indicates that AMR can help a model to better handle long-term dependencies by improving the entity recall.
In addition to utterance distance, we also consider word distance and observe a similar trend (as shown in Appendix \ref{fig:word_dis}).

\subsection{Case Study}
\label{sec:caseRE}
\begin{figure}[t]
    \centering
    \includegraphics[width=\columnwidth]{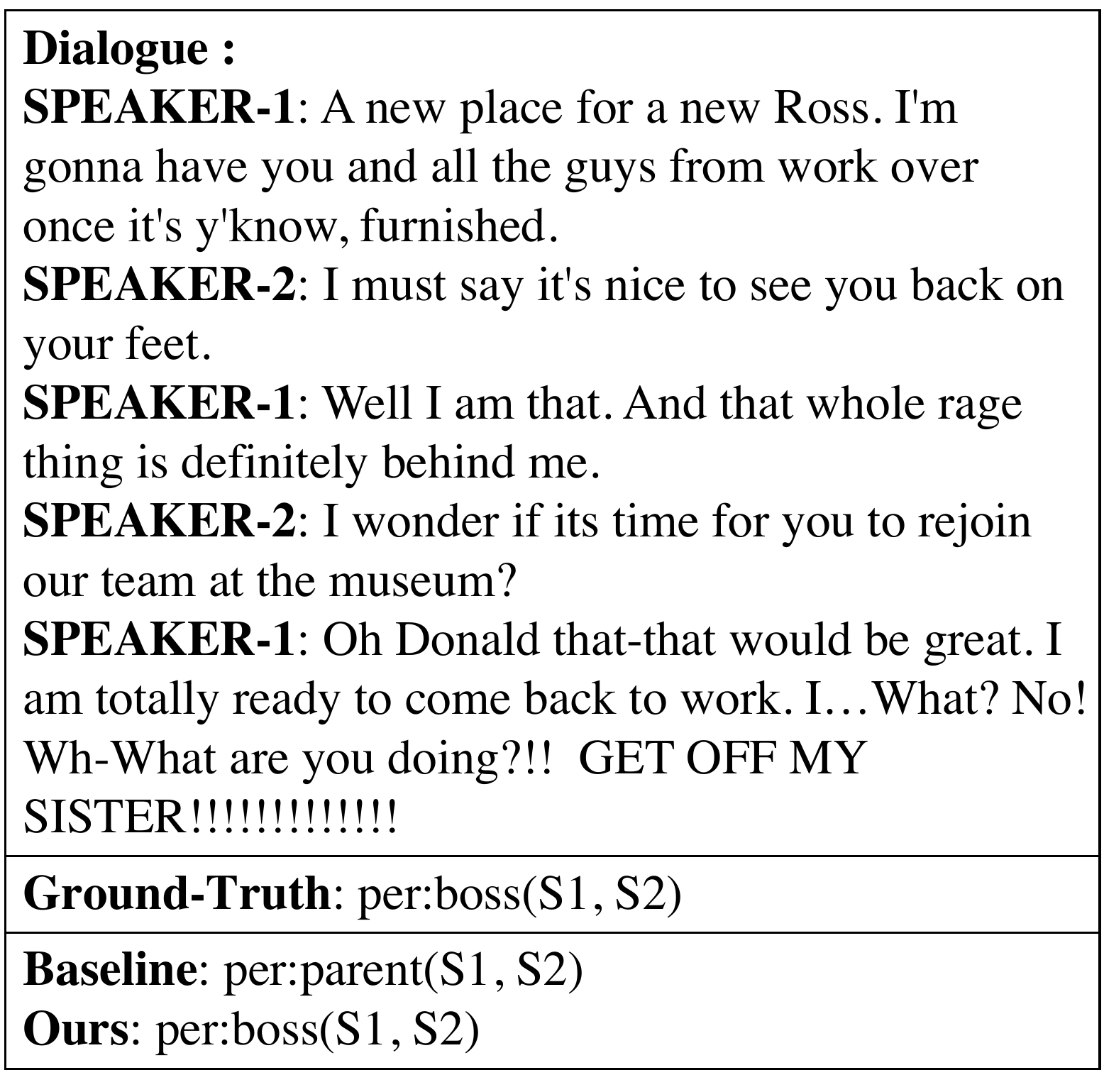}
    \caption{Case study for dialogue relation extraction.}
    \label{case_re}
\end{figure}

Figure~\ref{case_re} shows a conversation between a manager and an employee who might have taken a leave.
The baseline model incorrectly predicts that the relation between two interlocutors is parent and child.
It might be influenced by the last sentence in the conversation, assuming that it is a dialogue between family members.
However, the proposed model successful predicts the interlocutors' relation, suggesting it can extract global semantic information in the dialogue from a comprehensive perspective.

\section{Response Generation Experiments}
We conduct experiments on the DailyDialog
benchmark~\cite{li-etal-2017-dailydialog}, which contains 13,119 daily multi-turn conversations.
On average, the number of turns for each dialogue is 7.9, and each utterance has 14.6 tokens.
\subsection{Settings}

We take Transformer as a baseline.
Our hyperparameters are selected by word prediction accuracy on validation dataset.
The detailed hyperparameters are given in Appendix (See Table~\ref{tab:hyperparameters}).

\noindent\textbf{Metric} We set the decoding beam size as 5 and adopt BLEU-1/2/3/4~\cite{papineni-etal-2002-bleu} and Distinct-1/2~\cite{li-etal-2016-diversity} as automatic evaluation metrics. 
The former measures the n-gram overlap between generated response and the target response while the latter assesses the generation diversity, which is defined as the number of distinct uni- or bi-grams divided by the total amount of generated words.
In addition, we also conduct human evaluation.
Following~\citet{bao-etal-2020-plato}, we ask annotators who study linguistics to evaluate model outputs from four aspects, which are fluency, coherence, informativeness and overall performance.
The scores are in a scale of \{0, 1, 2\}.
The higher, the better.

\subsection{Automatic Evaluation Results}
\begin{table}
	\centering
	\small
	\begin{tabular}{lcc}
		\toprule
		\textbf{Model}&BLEU-1/2/3/4 &Distinct-1/2 \\
		\midrule
		Seq2Seq$^\dagger$ &33.6/26.8/-/- &3.0/12.8   \\
		iVAE$_{\texttt{MI}}$ &30.9/24.9/-/- &2.9/25.0  \\
		PLATO w/o L$^{\dagger\flat}$ &40.5/32.2/-/- &4.6/24.6 \\
		PLATO$^{\dagger\flat}$ &39.7/31.1/-/- &5.3/29.1 \\
		\midrule
		\rowcolor{mygray}
		Transformer &38.3/31.7/29.1/27.8 &5.8/30.5  \\
		\rowcolor{mygray}
        \texttt{Hier} &\textbf{41.3}/\textbf{35.4}/\textbf{33.2}/\textbf{32.1} &6.5/32.3 \\
		\rowcolor{mygray}
		\texttt{Dual} &40.8/35.0/32.7/31.5 &\textbf{6.6}/\textbf{33.0}\\
		\bottomrule
	\end{tabular}
	\caption{Performance on DailyDialog. Results marked with $\dagger$ are from~\citet{bao-etal-2020-plato}. Models marked with $\flat$ requires external corpus for pretraining.}
	\label{tab:mainDG}
\end{table}
Table~\ref{tab:mainDG} reports the performances of the previous state-of-the-art methods and proposed models on the DailyDialog testset.
For the previous methods, PLATO and PLATO w/o L are both Transformer models pre-trained on large-scale conversational data (8.3 million samples) and finetuned on DailyDialog. 
For completeness, we also report other systems including Seq2Seq~\cite{VinyalsL15} and iVAE$_\texttt{MI}$~\cite{FangLGDC19}.

Among the previous systems, PLATO and PLATO w/o L report the best performances.
Our Transformer baseline is highly competitive in terms of BLEU and Distinct scores. 
Compared with the Transformer baseline, both \texttt{Dual} and
\texttt{Hier} show better numbers regarding BLEU and Distinct,
and the gains of both models are significant ($p<0.01$).
This indicates that semantic information in AMR graphs is useful for dialogue response generation.
In particular, the gains come from better recall of the important entities and their relations in a dialogue history, which can leads to generating a more detailed response.


\subsection{Human Evaluation Results}
\begin{table}
	\centering
	\small
	\begin{tabular}{lcccc}
		\toprule
        \textbf{Model} & Fluency & Coherence  & Inf.  & Overall \\
		\midrule 	
		Transformer & 1.76 & 0.86 & 1.40 & 0.66  \\
		\texttt{Hier} & 1.86 & \textbf{1.04} &1.48 & 0.82 \\
		\texttt{Dual} & \textbf{1.88} &\textbf{1.04} & \textbf{1.52}&  \textbf{0.84} \\
		\bottomrule
	\end{tabular}
	\caption{Human evaluation results on DailyDialog. Inf. stands for Informativeness.}
	\label{tab:humanondaily}
\end{table}

We conduct human evaluation on randomly selected $50$ dialogues and corresponding generated responses of the baseline and our models.
As shown in Table~\ref{tab:humanondaily}, the Transformer baseline gives the lowest scores, while \texttt{Dual} sees the highest scores from all aspects.
Our main advantage is on the \emph{Coherence}, meaning that AMRs are effective on recalling important concepts and relations.
As the result, it makes it easier for our models to generate coherent replies.
Examples are shown in Figure \ref{case_daily} in Appendix.
Comparatively, all systems achieve high scores regarding \emph{Fluency}, suggesting that this aspect is not the current bottleneck for response generation.



\section{Analysis}
This section contains analysis concerning the effects of graph features, dialogue length and model robustness. 
We use \texttt{Dual} model for experiments since it gives slightly better results than \texttt{Hier}.
\subsection{Ablation on AMR graph}
\begin{table}
	\centering
	\small
	\begin{tabular}{lcc}
		\toprule
		\textbf{Setting} & DialogRE (v2) & DailyDialog \\
		\midrule
		Dialog-AMR(\texttt{Dual}) & \textbf{68.2} &\textbf{38.2/5.9}  \\
		\quad-Speaker &67.5 &37.7/5.7  \\
		\quad-Ident. concept &68.0 &37.9/5.8 \\
		\quad-Coref &67.8 &37.4/5.6 \\
		Utter-AMR &67.4 &36.9/5.6  \\
		Text &66.2 &35.4/5.5  \\
		\bottomrule
	\end{tabular}
	\caption{Ablation study on the development sets of both DialogRE (v2) and DailyDialog.}
	\label{tab:ablation}
\end{table}
Table~\ref{tab:ablation} shows the results of our best performing models on the two datasets regarding different configurations on the dialogue AMR graphs.
We report the average F1 score for DialogRE and the BLEU-1/Distinct-1 score for DailyDialog.
First, using utterance-level AMR improves the text baseline by 1.2 points and 1.5 points with regard to F1 and BLEU-1 scores, respectively.
This indicates that the semantic knowledge in formal AMR is helpful for dialogue modeling. 

Second, our manually added relations (in Section~\ref{sec:dialogue-AMR}) also leads to improvements, ranging from 0.5 to 1.0 in BLEU-1 score.
The speaker relation is the most important for dialogue relation extraction, a possible reason is that DialogRE dataset mainly focus on person entities. 
Also, co-reference relations help the most in dialogue response generation. 
The identical concept relations give least improvements among three relations. 
Finally, combining all relations to build a Dialog-AMR graph achieves best performance on both datasets. 

\subsection{Impact of Dialogue Length}
\begin{figure}
    \centering
    \small
    \includegraphics[width=0.95\columnwidth]{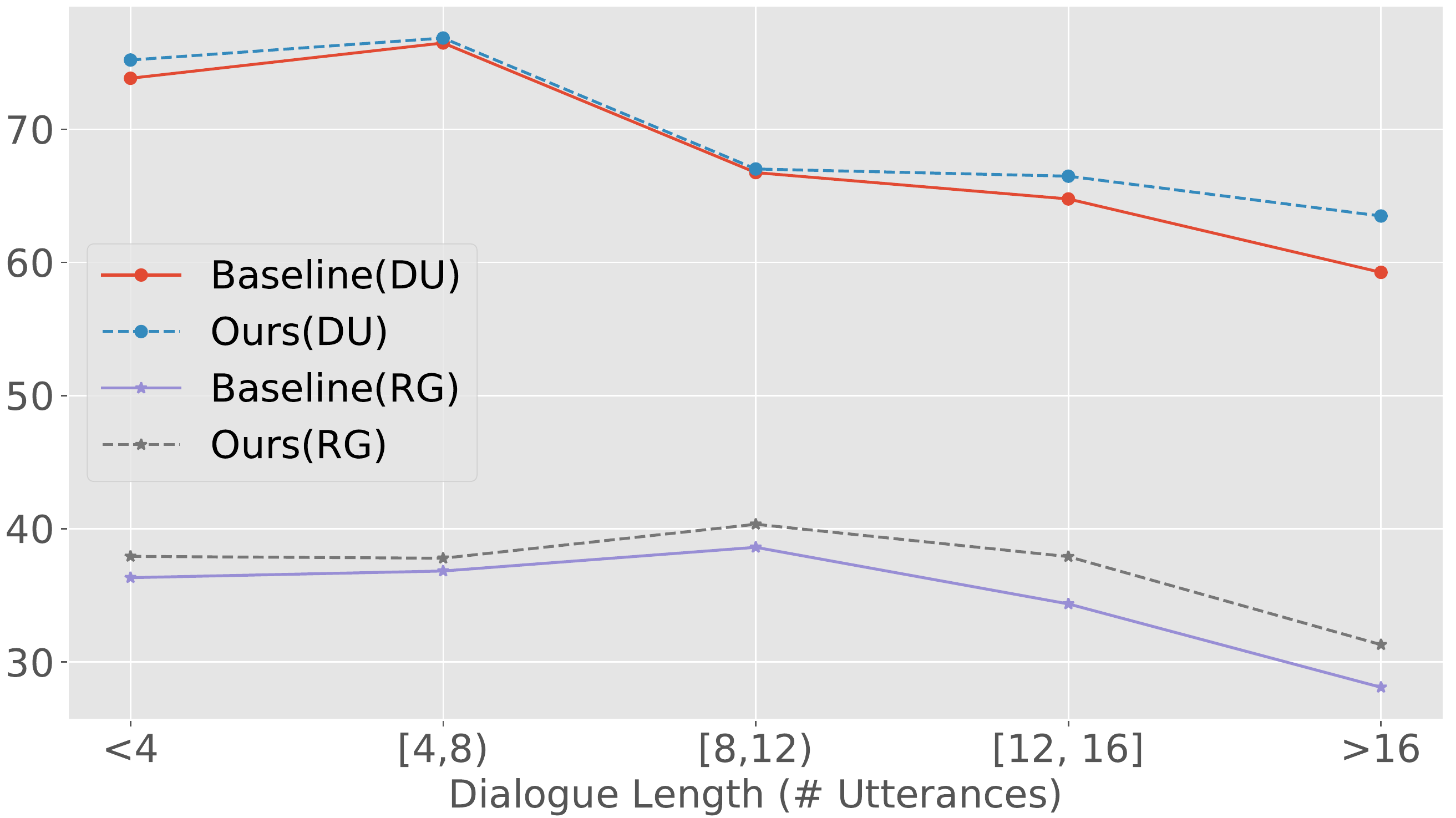}
    \caption{Devset performance against dialogue lengths.}
    \label{fig:dialog-len}
\end{figure}
We group the devset of DialogRE (v2) and DailyDialog into five groups according to the number of utterances in a dialogue.
Figure~\ref{fig:dialog-len} summarizes the performance of the baseline and the proposed model on dialogue understanding (DU) and response generation (RG) tasks.
In dialogue understanding, our model gives slightly better F1 scores than the baseline when a dialogue has smaller than 12 utterance. 
The performance improvement is more significant when modeling a long dialogue.
This confirms our motivation that AMR can help to understand long dialogues.
In dialogue response generation, our model consistently outperforms the Transformer baseline by a large margin on dialogues of different lengths, still with more improvements on larger dialogues.
Overall, these results are consistent with Table~\ref{tab:mainRE} and \ref{tab:mainDG}, showing that AMR can provide useful semantic information and alleviate the issue of long-range dependency.

\subsection{Robustness Against Input}
Recent studies show that neural network-based dialog models lack robustness~\cite{Shalyminov18,Einolghozati19}.
We select 100 instances from the testset of DialogRE (v2) where both baseline and our model gives true prediction, before paraphrasing the source dialogues manually (see appendix~\ref{sec:paraph} for paraphrasing guidelines.). 

Results on the paraphrased dataset are given in Table~\ref{tab:robustness}.
The performance of baseline model drop from 100 to 94.5 on paraphrased dataset. 
By contrast, the result of our model reaches 98.5, 4 points higher than baseline.
This confirms our assumption that AMR can reduce data sparsity, thus improve the robustness of neural models.

\begin{table}
	\centering
	\small
	\begin{tabular}{lcc}
		\toprule
		\textbf{Model} & Original & Paraphrased \\
		\midrule 	
		Baseline & 100 & 94.50 \\
		Ours & 100 & \textbf{98.50} \\
		\bottomrule
	\end{tabular}
	\caption{F1 on original and paraphrased testsets.}
	\label{tab:robustness}
\end{table}
\section{Related Work}
\paragraph{Semantic Parsing for Dialogue}
Some previous work builds domain-specified semantic schema for task-oriented dialogues.
For example, in the PEGASUS~\cite{zue-etal-1994-pegasus} system, a sentence is first transformed into a semantic frame and then used for travel planing.
~\citet{WirschingHuberKoelbletal2012} use semantic features to help a dialogue system perform certain database operations. 
~\citet{gupta-etal-2018-semantic-parsing} represent task-oriented conversations as semantic trees where intents and slots are tree nodes.
They solve intent classification and slot-filling task via semantic parsing.
\citet{ChengAABDFKKLPW20} design a rooted semantic graph that integrates domains, verbs, operators and slots in order to perform dialogue state tracking. 
All these structures are designed for specified task only. 
In contrast, we investigate a general semantic representation for the modeling of everyday conversations. 

\paragraph{Constructing AMRs beyond Sentence Level}
There are a few attempts to construct AMRs beyond the sentence level.
\citet{liu-etal-2015-toward} construct document-level AMRs by merging identical concepts of sentence-level AMRs for abstractive summerization, and \citet{liao2018abstract} further extend this approach to multi-document summerization.
\citet{o2018amr} manually annotate co-reference information across sentence AMRs.
We focus on creating conversation-level AMRs to facilitate information exchange more effectively for dialogue modeling.

\citet{bonial-etal-2020-dialogue} adapt AMRs on dialogues by enriching the standard AMR schema with dialogue acts, tense and aspect, and they construct a dataset consisting of 340 dialogue AMRs.
However, they propose theoretical changes in the schema for annotating AMRs, while we explore empirical solutions that leverage existing AMRs of the standard schema on dialogues.

\paragraph{AMR Parsing and Encoding}
Our work is also related to AMR parsing~\cite{flanigan-etal-2014-discriminative,KonstasIYCZ17,TitovL18,guo-lu-2018-better,zhang-etal-2019-amr,cai-lam-2020-amr} and AMR encoding~\cite{konstas2017neural,song2018graph,zhu2019modeling,song-etal-2020-structural,zhao-etal-2020-line,bai-etal-2020-online}.
The former task makes it possible to use automatically-generated AMRs for downstream applications, while the latter helps to effectively exploit structural information in AMRs.
In this work, we investigate AMRs for dialogue representation and combine AMRs with text for dialogue modeling.


\section{Conclusion}
We investigated the feasibility of using AMRs for dialogue modeling, describing an algorithm to construct dialogue-level AMRs automatically and exploiting two ways to incorporate AMRs into neural dialogue systems. 
Experiments on two benchmarks show advantages of using AMR semantic representations model on both dialogue understanding and dialogue response generation.

\section*{Acknowledgments}
Yue Zhang is the corresponding author. 
We would like to thank the anonymous reviewers for their insightful comments and Jinhao Jiang for his help for data preparation.
This work has been supported by Tencent AI Lab Rhino-Bird Focused Research Program.
It also receives support from the Westlake University and Bright Dream Joint Institute for Intelligent Robotics, and a research grant from Rxhui Inc.


\bibliographystyle{acl_natbib}
\bibliography{acl2021}

\clearpage
\appendix
\begin{table*}[!t]
    \centering
    \small
    \begin{tabular}{l|c|c|c}
        \toprule
        \multicolumn{2}{c}{\textbf{Setting}} & \textbf{DialogRE} &\textbf{DailyDialog} \\
        \midrule
        \multirow{7}{*}{Sequence Encoder} & Dropout & 0.1 &0.1 \\
        & Encoder Layers &12 & 4 \\
        & Attention Heads & 12 & 8\\
        & Embedding Size &768  & 512\\
        & Hidden Layer size &768 &512 \\
        & Word Vocabulary size &31k &16k \\
        & Feed-Forward Layer size & 3072 & 1024\\
        & Number of parameters &110M &38M \\
        \midrule
        \multirow{6}{*}{\shortstack{Graph Encoder\\(\texttt{Hier})}} & Dropout & 0.1 & 0.1\\
        & Encoder Layers &2 & 2\\
        & Attention Heads &8 & 8\\
        & Hidden Layer size &512 & 512\\
        & Relation Embedding size & 64 &64\\
        & Feed-Forward Layer size &1024 &1024\\
        & Number of parameters &4M &4M \\
        \midrule
        \multirow{7}{*}{\shortstack{Graph Encoder\\(\texttt{Dual})}} & Dropout & 0.1 &0.1\\
        & Encoder Layers &3 &4 \\
        & Attention Heads &8 & 8\\
        & Hidden Layer Size &512 &512\\
        & Relation Embedding Size & 64 &64\\
        & Concept Vocabulary Size &5.2k &10k \\
        & Feed-Forward Layer Size &1024 &1024\\
        & Number of parameters &11M &20M \\
        \midrule
        \multirow{7}{*}{\shortstack{Others}} & Optimizer & Adam &Adam\\
        & Batch Size & 48 &20\\
        & Learning Rate & 3e-5 &1e-4\\
        & Training Epoch & 30 &200\\
        & Decoder Layers &- &4 \\
        & Training Device &Tesla V100 &Tesla V100\\
        & Training Time & 120min & 48h\\
        \bottomrule
    \end{tabular}
    \caption{Hyperparameters of our models on DialogRE and DailyDialog.}
    \label{tab:hyperparameters}
\end{table*}
\begin{figure}[!t]
    \centering
    \includegraphics[width=0.95\columnwidth]{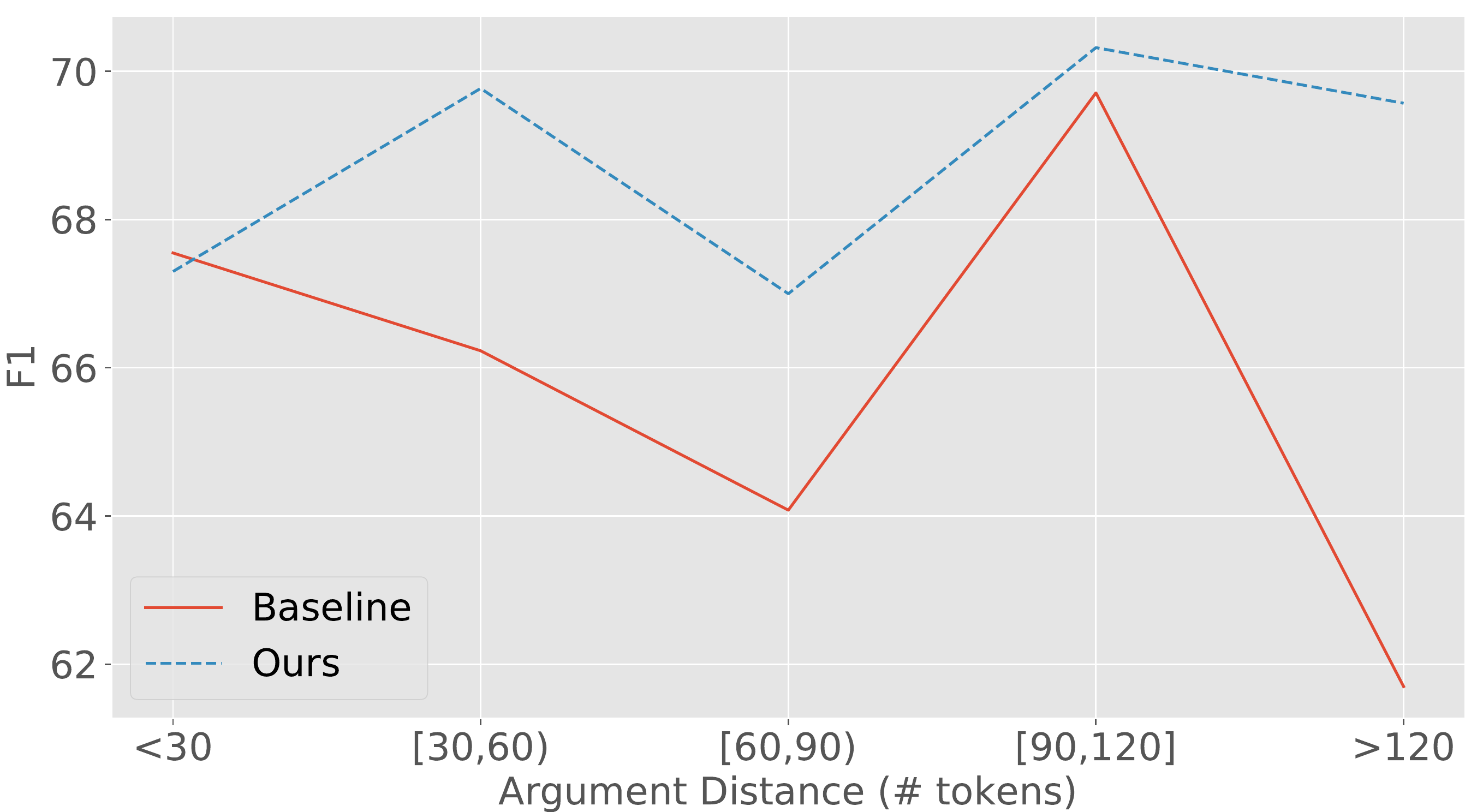}
    \caption{Performance against argument word distance.}
    \label{fig:word_dis}
\end{figure}

\section{Model parameters}
Table~\ref{tab:hyperparameters} lists all model hyperparameters used for experiments.
In particular, we share the word vocabulary of encoder and decoder for response generation.
We implement our baselines and proposed model based on Pytorch.
The preprocessed data and source code will be released at~\url{https://github.com/muyeby/AMR-Dialogue}.
\section{More Experimental Results}
\subsection{Impact of Argument Distance}
In addition to utterance distance used in Figure~\ref{fig:dis-utter}, we also consider word-based distance as a metric to measure argument distance.
Figure~\ref{fig:word_dis} shows F1 scores of baseline and our model on 5 groups of test instances. 
It can be seen that our model gives better results than baseline system among all distances longer than 30.
In particular, our model surpass baseline by 8 points when argument distance is longer than 120.
\begin{figure}[!ht]
    \centering
    \includegraphics[width=\columnwidth]{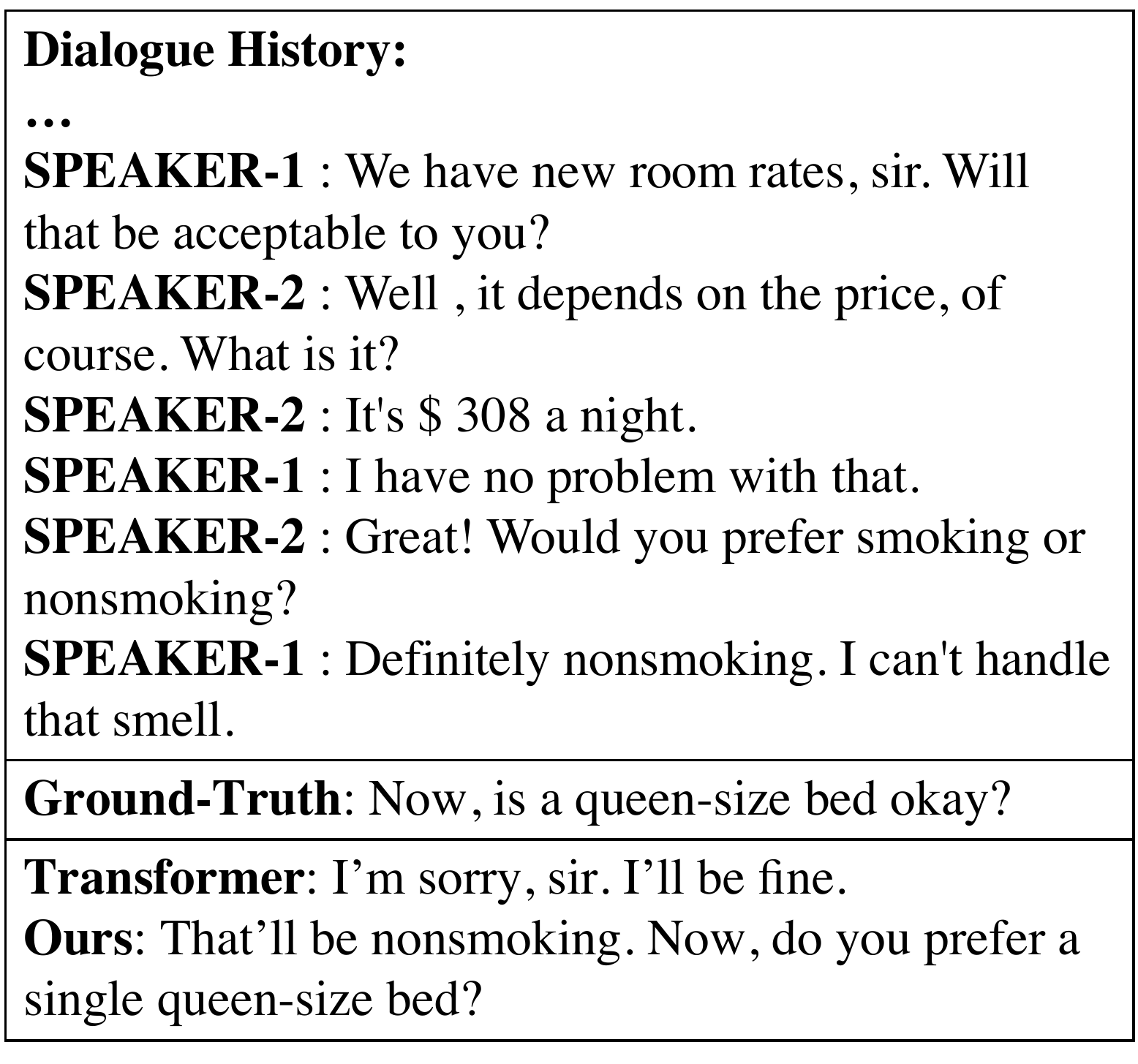}
    \caption{Case study for dialogue response generation.}
    \label{case_daily}
\end{figure}
\subsection{Case Study for Dialogue Response Generation}
\label{sec:caseDRG}
Figure~\ref{case_daily} represents a conversation between a hotel service and a guest who wants to book a room, along with its ground-truth response and model-generated responses.
We can observe that Transformer's output is general and not consistent with dialogue history.
While proposed models' outputs can capture the core information ``\textit{room}'' from the history, and are more relevant to the topic.
Besides, the output given by proposed model is semantically similar to the ground-truth output, but using novel words to response, indicating that the model not only captures the simple dependency between input and output sentences, but also learns deep semantic information of the dialogue history.
\subsection{Paraphrasing Guidelines}
\label{sec:paraph}
We ask annotators to paraphrase the dialogues following 3 guidelines: 

\textbullet~do not change the original meaning. 

\textbullet~paraphrase the sentence by using different lexicon and syntax structures. 

\textbullet~paraphrase the dialogue as much as they can.

We also ask a judge to evaluate whether the paraphrased dialogue (sentences) convey the same meaning of the original ones.
\end{document}